# A New Approach for Knowledge Generation Using Active Inference

Jamshid Ghasimi[a*]     Nazanin Movarraei[b†]

[a] Prof. Hessaby foundation, Tehran, Iran
[b] School of Mathematics, Institute for Research in Fundamental Sciences
(IPM), P.O. Box: 19395-5746, Tehran, Iran

**Abstract**

There are various models proposed on how knowledge is generated in the human brain including the semantic networks model. Although this model has been widely studied and even computational models are presented, but, due to various limits and inefficiencies in the generation of different types of knowledge, its application is limited to semantic knowledge because of has been formed according to semantic memory and declarative knowledge and has many limits in explaining various procedural and conditional knowledge. Given the importance of providing an appropriate model for knowledge generation, especially in the areas of improving human cognitive functions or building intelligent machines, improving existing models in knowledge generation or providing more comprehensive models is of great importance. In the current study, based on the free energy principle of the brain, is the researchers proposed a model for generating three types of declarative, procedural, and conditional knowledge. While explaining different types of knowledge, this model is capable to compute and generate concepts from stimuli based on probabilistic mathematics and the action-perception process (active inference). The proposed model is unsupervised learning that can update itself using a combination of different stimuli as a generative model can generate new concepts of unsupervised received stimuli. In this model, the active inference process is used in the generation of procedural and conditional knowledge and the perception process is used to generate declarative knowledge.

**Keywords:** semantic network, knowledge generation, free energy principle, active inference

## 1 Introduction

How knowledge is generated in the human brain is very important because of its effect on the study of cognitive features and functions such as problem-solving, judgment, or decision making [38, 30, 26]. At the same time, finding appropriate models for knowledge generation by the brain can pave the way for the realization of intelligent machines that, unsupervised, can generate knowledge and use it in processes related to their cognitive features and functional goals.

Human knowledge is formed based on the contents of its memory, therefore, given the function of memory and different features of knowledge, human knowledge can be classified into three categories: declarative knowledge, procedural knowledge, and conditional knowledge. Declarative knowledge is about what a phenomenon or object is, and procedural knowledge is about how things are done. Conditional knowledge leads to guiding a person in applying both declarative and procedural knowledge [1, 22, 32, 51].

*Email: jgh1001@yahoo.com
†Email: nazanin.movarraei@gmail.com

Various models for knowledge generation are mostly focused on one or at most two categories of knowledge. In addition, the process of knowledge generation in the existing proposed models is more descriptive and lacks computational steps or process analysis that can be converted into computational algorithms. At the same time, no model can address all three types of knowledge in an integrated manner. One of the most important and widely used models is the semantic networks model, which provides a model for generating semantic knowledge using the relationship between concepts. This model refers only to the generation of declarative knowledge and there is very limited information on procedural or conditional knowledge generation and expression. The model of semantic networks is based on semantic memory and despite the advances that have been made to compute it; it has many limitations in providing a computational model for generating procedural or conditional knowledge.

The current paper presents a model for the generation of declarative knowledge that simultaneously supports, analyzes, and generates procedural and conditional knowledge computationally. The proposed model takes into account concepts, the relationship between concepts, and the features or stimuli of concepts generators. Concepts or objects are the smallest unit of declarative knowledge formation in a semantic network structure or knowledge generation. This model is based on the free energy principle (FEP) in the brain, which shows the process of generating concepts from stimuli [5, 22]. For this purpose, while reviewing the FEP model and examining its variables, the model has been adapted in a way that can show how to generate different types of knowledge. Finally, a model is presented that, while being able to compute different types of knowledge, can also show the process of knowledge generation. This model shows the process of learning and updating knowledge and generating concepts under different conditions such as meaning-making using stimuli, executive and procedural functions, as well as the acquisition of executive skills [19]. In terms of computation or modeling, this model uses the inference attribute of the human mind to generate concepts by receiving sensory stimuli [23].

## 2  Semantic memory and the formation of semantic

Declarative knowledge is the result of the data in semantic memory processed by the brain [11, 43]. Perception of concepts, and their stimuli and the relationship between different concepts based on the commonalities in their stimuli form declarative knowledge. Stimuli, because of the features of concepts, generate new concepts that are then stored in memory. It should be noted that in semantic memory, the meaning of a concept is preserved and not the word itself or the grammatical features of a conceptual sentence. Concepts are generated through their stimuli. In other words, concepts are hidden variables in the environment that can be discovered or inferred through stimuli.

In addition, declarative knowledge can be presented orally (linguistically) and in writing. In this case, declarative knowledge is generated by information that is constructed as a hierarchy in the semantic memory and the form of a network consisting of concepts and their relationships (semantic networks) [31].

Each concept could have several stimuli. The stimuli of each concept can be varied and abundant, which depending on the observer may differ in quality and number, however, most concepts and objects in the environment create stimuli that affect the observer factors equally. Networks of interconnected concepts are constructed about the semantic networks, which may differ depending on the type of culture, or the capabilities of agents (observers).

Thus, semantic knowledge is the result of input perceptions and inference of these perceptions, which is stored in semantic memory as a communication network. This network creates semantic organization, which allows the agent to retrieve information. This type of semantic network is in a hierarchical manner and a combination of concepts in the form of nodes and relationships between concepts. A graphical representation of a semantic network is shown in Figure 1 [4] relation between concepts is represented by arrows. In Figure 1 semantic network, Is A means a direct link between two concepts, for example, "stork is a bird" or Capable of means the ability of one concept to perform another.

Semantic networks are presented in different forms or models with similarities and differences [33, 29, 47, 50] but the common feature of all is the on concepts and the stimuli of concepts to separate or categorize concepts. These models include [12, 40]:

- Feature comparison model: Concepts are categorized by matching their stimuli, and according to the stimuli of the target group.
- Hierarchical network model: Concepts are organized by using hierarchical relationships in the semantic network.
- Spreading activation model: This model emphasizes interconnected concepts. Concepts once activated in the mind, expand along network paths and activate other concepts, however, the farther away from the original concept, the less active the subsequent concepts become. This model shows a clear picture of the semantic relationships between concepts that decrease over time.

Semantic network models, despite their many strengths in explaining the memory functions that lead to knowledge generation, also have weaknesses. For some concepts, the hierarchical model may not find a feature that puts it at one of the levels of the hierarchy, or it may even have some kind of contradiction in predicting the levels of the hierarchy for some concepts. In addition, this model neglects the typical effect. The identifier effect states that members who are more identifiable at one level are easier to categorize than less representative members are. The success of the scalable activation model also significantly depends on the proper explanation of the test results; otherwise, it is considered an inefficient model, although it well demonstrates the ability of semantics preparation. Semantic network models are so flexible that sometimes this flexibility makes them impossible to predict. The process of knowledge generation in semantic networks is the result of stimuli generation and the relationship of received concepts with stimuli. This feature of semantic networks represents a Markovian model [52, 2] that has the following two basic features:

- Concepts are not fully known, understanding or inferring each does not mean inferring all existing concepts, and they are known as hidden variables that must be inferred through stimuli. So that in terms of the agent, some concepts can be understood through inference, and some are hidden in the environment.
- Each concept is deduced under a probabilistic process through its stimuli generated by that concept. Given the models of semantic networks, the way of generating semantic knowledge is the result of converting continuous perceptual signals into discrete concepts using the Markovian model.

# 3 The relationship between concepts and stimuli

In an interaction between man and the environment and the connection between the two, each concept is shaped by its stimuli. Depending on whether the concepts are sensory or associative, their properties are perceived by the agent through sensory perceptions or other data such as verbal or written data. As soon as the environmental stimuli are received by the brain, either the previous concepts are activated and primed, or a new concept will be generated. For example, when an agent hears a dog barking while simultaneously observes the shape of a dog, the concept of a dog comes to mind. In this case, sound and image, which are received through the two sensory modalities of phonology and vision, will be formed as stimuli and the dog itself as a concept [15, 14]. Each concept is perceived as a combination of stimuli. Several sensory or abstract stimuli can be considered that are located in the brain and memory for priming several concepts. Dogs, for example, have an objective meaning derived from sensory stimuli, while the word justice has an abstract meaning [14]. Of course, concepts can have both objective and abstract properties, such as the word home.

If n is several concepts, then m stimuli can be considered for all of them. Some of these stimuli may be common to different concepts; synonyms are formed if several concepts share all of their stimuli. The concepts and the stimuli sets have been shown as follows:

$$\text{Concepts set}: S = \{s_1, ..., s_n\}$$

$$\text{Stimuli set}: R = \{r_1, ..., r_m\}$$

Where n is the number of concepts and m is the number of stimuli. The relationship between concepts and stimuli can be considered as a matrix with m × n dimensions (A matrix), which is also called the matrix or model for generating concepts from stimuli. The rows of this matrix are the number of stimuli and the columns are the number of concepts. In the simplest case, the components of the A matrix ($a_{ij}$) will be binary. In this case, if a particular stimulus, like stimulus j, does not exist in a particular concept, like i, the component $a_{ij}$ is zero, otherwise, it is equal to one. In a communication system consisting of a transmitter (could be the environment or a human agent or a computer) and a receiver (could be a computer or a human agent), the amount of energy required to transmit information is obtained from Equation 1 [14]:

$$\Omega(\lambda) = -\lambda I(S, R) + (1 - \lambda) H(S) \qquad (1)$$

Where, $\Omega(\lambda)$, $I(S, R)$ and $H(S)$ are the energy function, the information transfer between stimulus and the concepts, and the entropy of the concepts, respectively. $\lambda$ is a parameter of controlling the balance between information transfer maximization and cost minimization of communication (entropy of concepts), that is in the range $\lambda \in [0, 1]$.
The stimulus ( R set) is the information that is sent by the environment or person to another person or machine to convey concepts. Here, we consider the recipient to be a human being who must be able to deduce concepts in the environment by receiving environmental stimuli. Of course, if the sender is a human being who wants to transfer information to the recipient (human agent) through speech or writing, according to Zipf's Law [45, 52], the information transfer energy is minimized.
In this case, both the transmitter and the receiver (speaker and listener) act to minimize free en-ergy, which according to the principle of least effort try to maximize information and minimize the entropy of concepts (cost). According to the equation of information transfer energy, this process should be done to minimize $\Omega(\lambda)$ energy , in which case the two terms of information and entropy are effective. If the entropy increases, in practice, the amount of energy also increases, while the amount of information must be maximized. It means that to minimize the energy, the information $I(S, R)$ , and the entropy $H(S)$ must work in two opposite directions. Hence, we need to find an optimal point concerning $\lambda$ to achieve energy minimization concerning information maximization of both concepts and stimuli and minimization of entropy. For $\lambda \approx 0.41$ [14, 53], the system is very close to this condition. So that, the conflict between the minimum entropy of concepts and the maximum of information is resolved for the appropriate value $\lambda$ . In this case, the choice $\lambda$ can be considered as a policy choice to predict energy consumption. A Matrix, which shows the relationship between concepts and stimuli, is a concept generative matrix of stimuli that works under the policy of $\lambda$ choice to minimize the energy of information transmission, according to the received stimuli. Thus, we will have a generative model of concepts based on stimuli.

If there was a smart machine on the transmitter side instead of the human agent, it can be assumed that the entropy of the transmitted information would be zero [14]. That is, the transmission energy is as follows:

$$\Omega_o = I(S, R) \qquad (2)$$

In this case, the minimizing effort will be only on the receiver side. In addition, if instead of the transmitter, there is an environment in which the human agent is located, then we will have a one-sided minimum effort on the side of the receiver (human agent) to receive concepts from the environment. According to Zipfs law, if the sender is a human, most entropy changes result from the use of synonymous words or concepts that give rise to similar stimuli, as well as the use of words with high frequency and low variation and words with low frequency but a high variation. If only a small number of synonyms are more frequent, there will be a high cognitive cost. Also, if all the concepts in the text sent by the sender have an equal probability of sending, the entropy value $H(S)$ will be maximized, and if only one concept is submitted, this entropy will be zero. The amount of combined information is shown in Equation 3 [16]:

$$I(S, R) = \sum_{i=1}^{n} \sum_{j=1}^{m} p(s_i, r_j) \log \frac{p(s_i, r_j)}{p(s_i)p(r_j)} \qquad (3)$$

$p(s_i)$ , $p(r_j)$ and $p(s_i, r_j)$ are the distribution density of concepts, the distribution density of stimuli, and the distribution density of concepts-stimuli respectively. Equation 3 shows that the combined information of concepts-stimuli is equal to the amount of divergence between the possible distributions of concepts-stimuli with multiplied by the distributions of concepts and stimuli or the KullbackLeibler divergence [15] and we have:

$$I(S, R) = D_{KL}(p(s_i, r_j) \| p(s_i)p(r_j)) \qquad (4)$$

Therefore, the information transfer energy function can be as Equation 5:

$$\begin{cases} \Omega(\lambda)/\lambda = -D_{KL}(p(s_i,r_j)\|p(s_i)p(r_j)) + a_\Omega(H(S)) \\ a_\Omega = \frac{1}{\lambda} - 1 \qquad\qquad , 0 \leq \lambda \leq 1, \quad 0 \leq a_\Omega \leq \infty \end{cases} \quad (5)$$

The relationship between concepts and stimuli is a many-to-many relationship, meaning that each concept (or each word) may have different stimuli, and on the other hand, there might be several concepts for each stimulus. In this case, if stimuli jointly refer to different concepts, those concepts form a synset.

The $p(s_i, r_j)$ matrix, as a generative model and for each possibility of $s_i, r_j$, will be a method to generate concepts through stimuli. In such a model, with the introduction of new stimuli or a combination of existing stimuli that cannot refer to any of the concepts, one can expect the generation of new concepts and thus the expansion of the generative model matrix ( $p(s_i, r_j)$ matrix or A). The introduction of new concepts means an increase in the entropy of $H(S)$ that the receiver (human agent) can update his model by reducing this entropy and thus minimizing the information transmission energy.

This matrix model is another representation of semantic networks that allows the generation of semantic or declarative knowledge (receiving and expanding concepts and their relationship with each other). This model of semantic networks has the computational ability as well as present the process of generation and introduction of new concepts into the semantic network compared to the hierarchical models or the model of expanding activation or dendrogram of concepts with separation of sensory modalities or applied effects of concepts. Also, how the concepts are related and their similarity [49] can be shown in the form of a concept overlap matrix. This matrix is the result of the generation A matrix. Each of the $s_i$ concepts can be represented as a combination of $r_j$ stimuli [7] such as Equation 6:

$$s_i = (a_{i1}, a_{i2}, ..., a_{im}) \qquad (6)$$

$a_{ij}$ is the component of the A matrix, and $s_i$, is the i-th concept represented by the vector of the components related to the stimuli in the A matrix.

To calculate the similarity of concepts $s_i$ and $s_j$, one has:

$$\cos\theta_{ij} = \frac{\sum_{k=1}^{m} a_{ik} a_{jk}}{\|s_i\| \cdot \|s_j\|} \qquad (7)$$

$Where$, $\cos\theta_{ij}$ shows the degree of similarity between the $s_i$, $s_j$ concepts. Given that this value is in the range [0,1], the degree of similarity of the two concepts can be shown as probabilistic or as a percentage. So the similarity matrix will be a $n \times n$ matrix with components between zero and one. The process of generating a concept from stimuli is a Markov process in which hidden variables (concepts) are inferred by stimuli.

## 4 Free energy principle in the brain

According to the FEP, the brain is the existence of hypothesis tested [13]. It is continuously updating its hypotheses based on the concepts it acquires from perceptions received from the environment so that these concepts can be inferred by generating different sensory stimuli and influencing sensory perceptions [22] based on Bayesian inference and in a process as Markov process [46, 52]. The brain will be disturbed if it is unable to find out the truth and make predictions for the stimuli received through sensory inputs such as visionary or aural. This property of the brain is explained based on entropy properties and analogy to the principle of thermodynamic free energy [21].

The entropy of the brain increases as the agent receives new sensory inputs. If this increase remains the same or increases over time, it can lead to cognitive impairment by the brain. To prevent such a situation, it is necessary to reduce the generated entropy rapidly. In a thermodynamic system, entropy reduction occurs through the consumption of free energy [17]. Thermodynamically, the brain acts as an open system, so it can receive energy from the outside. Also, due to its self-organization nature [3] the brain can use its free energy to reduce entropy.

Receiving sensory data and increasing entropy causes a surprise, because of inconsistency of the received data with the hypothesis in the brain, which should be eliminated immediately through the consumption of free energy [18]. Accordingly, the FEP states that any adaptive change in the brain will minimize free energy. This principle is a logical analogy of the thermodynamic free energy model based on complex implications and an attempt to describe the structure and function of the brain. In this case, with the receipt of sensory data and the increase of surprisal, the synaptic connections of the brain are updated through coding [17].

If the variable is as sensory inputs, the entropy of these inputs is obtained from Equation 8:

$$H(\varphi) = -\int p(\varphi|m) ln p(\varphi|m) d\varphi$$
$$= \lim_{T \to \infty} \frac{1}{T} \int_0^T -ln p(\varphi|m) dt \qquad (8)$$

*Where*, $H(\varphi)$ is the entropy of sensory stimuli, m is the generative model that generates concepts from sensory stimuli, and t is a time variable. $p(\varphi|m)$ is the probability of sensory stimuli and the amount of surprisal (or the amount of self-information) is negative of $ln p(\varphi|m)$. Studies show that the brain uses Bayesian inference to induce sensory stimuli, i.e. environmental concepts to prevent the effect of surprise [27]. Equation 8 shows that entropy minimization is associated with surprisal compression over time. In this way, the brain, based on the structure of its hypothesis tested, is constantly engaged in minimizing its predictions errors based on these hypotheses. In other words, free energy is the same as prediction error.

In the FEP of the brain, two terms of surprise and the divergence between the estimated inferential posterior probability, $q(\theta)$ and the real probability model, $p(\theta|\varphi)$, determine the amount of free energy, in which $\theta$ the same concepts, or environmental states, generate sensory stimuli [18, 28].

Given the amount of surprisal and their effect on free energy, the surprisal bound is determined with the $-\log p(\varphi)$ term, which after the update; one will have a model of how surprise is generated in exchange for sensory perceptions. Given that any adaptive change in the brain minimizes free energy, this minimization takes place through two processes of perception and action [17, 18]:

- Perception means changing expectations to reduce entropy and prediction error,
- Action means changing the configuration by affecting the biological agent on the environment to change sensory stimuli and to avoid surprisal.

By combining perception and action, it is possible to adapt to new sensory stimuli. This pro-cess is called active inference.

The equations of free energy in the two processes of perception and action will be in the form of equations 9.

$$\text{Perception to optimize the bound} \begin{cases} F = Divergence + Surprise = D_{KL}[q(\theta|\mu)\|p(\theta|\varphi)] - lnp(\varphi) \\ \\ \mu = \arg\min_{\mu} Divergence \end{cases}$$

(9)

$$\text{Action to minimize the bound on surprise} \begin{cases} F = Complexity - Accuracy = D_{KL}(q\|p(\theta)) - <lnp(\varphi(a)|\theta,m)>_q \\ \\ a = \arg\max_{a} Accuracy \end{cases}$$

In these equations, also $\mu$ is the internal and coded concepts of the model in the brain and a is the actions required to influence sensory stimuli in order to reduce free energy.

The $\mu$ will change as the model is updated. Hidden environmental concepts are parametrized by the internal states of the brain. It is necessary to adopt policies, $\pi$ (action choice), to select appropriate actions, to minimize free energy [36, 39]. According to new sensory perceptions, these actions can change discontinuously in time steps.

It means actions will be a function of the policies in question at each time step($\tau$). The set of active inference process variables based on selected policies and actions at different stages is given in Table 1 [22, 24]

Figure 2 shows that the agent, instead of inferring the causes of new sensory stimuli, must perform an action that best adapts the stimulus to its generative concepts (the same mechanism of active inference). A very important assumption in this process is that the hidden concepts in the environment may not be received directly, but are inferred as a Bayesian process by receiving the sensory stimuli they create. If there is no action to receive stimuli and influence the environment, the prediction made about the concepts is the same as Bayesian inference, otherwise, inference will be active inference.

The actions selected under different policies can compress the error of predicting sensory stimuli, in which case the brain encodes a probabilistic relationship among internal states, sensory stimuli, behavioral responses, and outputs of these behaviors. This model shows that the brain not only estimates the closest similarities, but also the distribution of variables related to stimuli and concepts. The minimization of free energy is done according to the two functions of recognition probability density, namely $q(\theta)$ and generative density $p(\theta, \varphi)$ [6, 8]. The joint density function $p(\theta, \varphi)$ represents how concepts are generated concerning sensory stimuli [25]. This function is usually obtained by using the Bayesian formula and the density function of similarity and $p(\varphi|\theta)$ marginal distribution ($\theta$) shown in Equation 10.

$$p(\theta, \varphi) = p(\theta|\varphi)p(\varphi) = p(\varphi|\theta)p(\theta) \qquad (10)$$

Given that the agent needs hypotheses that represent the generation of concepts concerning sensory stimuli, the presence of recognition density and generative density functions will be necessary to minimize free energy. With the new sensory stimuli, the generative model (generative density) is updated and can generate new hidden variables (environmental states or concepts) according to these sensory stimuli. In other words, we will have a generative model that, unsupervised, can generate new and credible hidden states based on previous distributions and new data (sensory stimuli). In this case, the brain, under the mechanism of active inference, selects sensory stimuli according to its previous beliefs and minimizes the complexity of their representation, and by minimizing free energy, reaches a steady state of imbalance [8]. To minimize free energy, the prediction error is minimized, while at the same time the conditional density entropy, $H[q(\varphi|\mu)]$ is maximized. This is the same conflict in the stimulus-concept model of semantic networks. In this case, if the entropy of the sensory stimuli $H(\varphi|\mu)$ is minimized by minimizing the free energy over time, the conditional entropy $H[q(\varphi|\mu)]$ must be maximized.

At any given time, this phenomenon is due to the need for a balance between complexity and accuracy. Perception, action, and policy will regulate the free energy minimization mechanism [25, 44].

According to Equation 9, the operations that are selected according to different policies are necessary to minimize the divergence between the two functions of recognition density $q(\theta)$ and the function of the true posterior conditional probability $p(\theta|\varphi)$ [6]. On the environmental side, one will have the correct and definite concepts ($\theta^*$) the density of stimulus generation, i.e. , $p(\varphi)$ and the inferential generation model $p(\theta|\varphi)$. $\theta^*$ are real concepts without considering the inference of the agent, while, $\theta$ are concepts that are inferred by the agent and through observations (sensory stimuli). In this case, the conditional probability of producing sensory stimuli $p(\varphi|\theta)$ is due to the agent's previous beliefs about environmental concepts or states [31]. Active inference and free energy minimization evolve the model of generating concepts over time, maximizing the evidence for observations (stimuli). Introduces a model of free energy of the brain, which, like semantic networks, can generate semantic knowledge, and at the same time can also generate procedural knowledge by performing actions on the environment. This free energy model of the brain is called the FEP knowledge generation model.

## 5  Knowledge generation based on the FEP

The brain FEP model shown in Figure 2 introduces how concepts are generated concerning receiving new sensory stimuli. This process can be extended to the model of knowledge generation. In addition to declarative knowledge, it also includes procedural and conditional knowledge. This model is shown in Figure 3.

Figure 3 shows two different loops I and II. Loop I includes prediction, perception, and brain error. If the inference is passive (environment is not affected by agent), loop I is the brain's only functional process on sensory stimuli. The prediction error changes the predictions through perception, resulting in a change in the brain's internal coding and updating. Thus, if there is no action to change the sensory inputs, Bayesian-type learning and inference is done by updating previous probabilities (previous beliefs) concerning receiving new sensory stimuli. This process is similar to the computational model of updating semantic networks in the generation of declarative knowledge. The learning in the FEP of the brain means detecting and reducing the prediction error in loop I [19, 48]. Otherwise and in case of the need to influence sensory stimuli that have faced surprisal and to minimize free energy and avoid surprisal in the future, loop II is enabled which includes a predictive error, policy selection, and appropriate action and impact on the environment (sensory stimuli). In this case, Bayesian probabilistic inferences become Bayesian variational inferences, due to the need to use different policies [34] and eventually the inference will be activated. This feature of the model, which uses predicted actions to minimize the free energy [1, 6] means learning the concepts that under different actions, by receiving sensory stimuli, generate procedural knowledge. For this case, it is necessary to learn and update the parameters of the brain, in combination with the loop I. Thus, if the process of learning and updating the brain causes the agent to automatically learn a series of actions that minimize free energy, this type of learning leads to the generation of procedural knowledge.

After the generation of procedural knowledge, loop I will be inactive (without affecting the updating and coding) and to use loop II the agent performs its automatic activities by using memory and brain coding after updating and generating procedural knowledge. If both loops are active the agent can perform the desired actions on the environment (usually automatically) simultaneously with the extraction and application of concepts from semantic memory that are already under the process of active inference, perceived and updated them in their brain, it means the generation of conditional knowledge. Accordingly, the knowledge generation process for three examples is given in Table 2.

The brain function in a FEP based knowledge generation will be as follows:

- According to previous stimuli and hypotheses, the brain infers a set of concepts and considers a model of the inferential generation that predicts the future by receiving sensory stimuli.

- By receiving new sensory stimuli, if these stimuli are not able to infer previous concepts, the brain generates new concepts in return for these stimuli. In other words, if a stimulus or a set of several stimuli does not engage with any of the concepts, it is necessary to add a new hidden concept to the model [48].

- It is possible to reduce the surprisal of receiving stimuli by influencing the factor on sensory stimuli. In this way, the process of active inference is performed based on perception and action, and the brain updates the model of generating concepts from stimuli by changing and correcting future predictions.

In this case, the brain re-encodes a kind of probabilistic relationship between stimuli and concepts, which is a mapping of the generation model in the brain. Thus, using the principle of free energy and the processes of perception, action, and learning (updating the internal parameters of the brain), inferences and the construction of a hypothesis or model are integrated to generate different knowledge. In all these stages, concepts as environmental states are hidden variables that are inferred indirectly through sensory stimuli.

- The proposed knowledge generation model is a discontinuous model of concepts and sensory perceptions that due to the abstraction of concepts in a related set, one will have a discontinuous model of concepts. By learning or inferring concepts, the connected spaces of sensory stimuli are mapped into discrete entities called concepts. Abstraction of concepts provides the power to easily generalize and transfer knowledge among agents.

Given that the distribution of stimuli or concepts is not clear from the beginning, the Bayesian nonparametric (BNP) method is used for the learning process [2]. In an unsupervised manner, this method helps to upgrade and update the model, by increasing the number of observations (stimuli), over time, with high flexibility. Categorical distribution ($Cat$) is used to demonstrate the distribution of concepts, stimuli, and the possible distribution of policies ($q(\pi)$) because of the abstraction of these variables. Given that the Dirichlet distribution ($Dir$) is a conjugate prior of polynomial distribution, it can be appropriately used in Bayesian inference. The polynomial dis-tributions and the Dirichlet process are commonly used in nonparametric statistics and the use of discontinuous variables such as concepts. Accordingly, the variables and functions of the concept generation model are given in Table 3 [9, 10].

In Table 3, G is the expected free energy (EFE) based on the selected policy. represents the softmax function, which transmits the EFE values to the probability range [0,1] to determine the probability of selective policies ($p(\pi)$) per policy. The process of generating concepts and showing the impact of variables on the generative model can be seen in Figure 4 [37, 35].

Given the relationships and equations in Table 3 and Figure 4, one has a Bayesian decision-making process in which the EFE, G, determines the policies needed to understand the concepts in the future (to minimize free energy). There are also concepts based on the previous initial distribution function and the D matrix, for the agent to be believable that as the process continues, and as new sensory stimuli are received, the concepts are updated in a Bayesian inference under the conversion B matrix. At the same time, Figure 4 shows how to generate concepts through sensory stimuli by A matrix.

The probabilistic generation model is represented in Equation 11 [51, 50].

$$p(\tilde{\varphi}, \tilde{\theta}, \pi) = p(\theta_1)p(\pi) \prod_\tau p(\theta_{\tau+1}|\theta_\tau, \pi)p(\varphi_\tau|\theta_\tau) \qquad (11)$$

According to the process of learning and updating the generative model, the equations of perception, planning and action are shown in Table 4.

# 6 Comparison of knowledge generation models through semantic networks and the principle FEP of the brain

Learning and knowledge generation in both of semantic networks model and FEP of the brain is done by minimizing free energy. In semantic networks and the process of data transfer through stimuli to humans, the generation of concepts (meaning) is done by energy minimizing (Equation 1), and in the FEP of the brain, it is done by minimizing variable free energy which is dependent on human perception and action ( Equations 9). The comparison of these two approaches in knowledge generation is as follows:
- In the semantic network model, the inference is based on Bayesian probabilities and inferences, while in the FEP model, the inferences are based on the Bayesian variability and happen in the perception-action process.
- In the semantic network model, only the declarative knowledge is generated, but in the FEP model, all three types of declarative, procedural, and conditional knowledge are generated.
- Both models act as generative model to generate concepts of stimuli.
- Updating it over time is possible in both models.
- In the model of semantic networks, to separate concepts, the combination of stimuli is used to generate concepts, while the FEP uses the difference between prediction and sensory perceptions and the effect of the factor on sensory stimuli.
- Both models correspond to the characteristics of a Markov model in the process of learning and inference.
- Elimination of loop II in the FEP (Figure 3) eliminates the generation of procedural knowledge and active inference and turns it into a declarative knowledge generation model similar to the semantic network model.

- Perception is an important part of the FEP model in such a way that, while the semantic networks and free energy of information transmission models does not have such possibility, a computational model can be presented for human perception when receiving sensory stimuli.
- The semantic relationship between concepts in the model of semantic networks is determined based on the matrix of similarity or Euclidean distance of concepts [11, 41]

# 7 Conclusion

This article examined how knowledge is generated in the brain and proposed a model for it. The model of semantic networks is widely considered in the generation of declarative knowledge, but it is not possible to explain in detail the process of procedural and conditional knowledge generation. By classifying it into two loops of declarative knowledge generator and procedural knowledge generator, a model was proposed for generating different types of knowledge using the FEP model of the brain, which shows the ability to learn concepts in the human brain. This model, while calculating the method of generating different types of knowledge in the brain, can be very flexible in updating the model and generating concepts of sensory stimuli in a perceptual-action process. In the continuation of this research, while addressing sensory stimuli that are involved in the process of generating concepts, semantic or abstract stimuli can also be considered in the brain, which can generate new concepts without sensory stimuli. The application of this model can be examined in specific examples, such as knowledge generation in intelligent machines. Also, in the case of cognitive diseases, especially Alzheimer's, probability it is possible to control the disease based on the process of producing semantic or procedural knowledge.

# 8 Acknowledgment

# 9 Tables and Figures

| Description | Expression |
|---|---|
| Sensory stimulus variable | $\varphi$ |
| Hidden concepts variable | $\theta$ |
| Specific action on the time scale, $\tau$ | $a_\tau = \pi(\tau)$ |
| Total time steps | $T$ |
| Time steps from the first step to the $T$ step | $\tau \in [1, 2, ..., T]$ |
| Selective policy at each time step for a specific action | $\pi$ |
| A set of policies related to each action at each time step | $\pi_\pi = (a_{\pi,1}, a_{\pi,2}, ... a_{\pi,T})$ |
| A set of sensory stimuli at all of the time steps | $\tilde{\varphi} = [\varphi_1, ..., \varphi_T]$ |
| A set of hidden concepts at all of the time steps | $\tilde{\theta} = [\theta_1, ..., \theta_T]$ |
| Coding concepts in the brain | $\mu$ |
| Variable Free Energy | **F** |

Table 1: Active Inference model variables

| Example | Knowledge type | Active loop for learning and knowledge generation | Active loop to use knowledge | Inference |
|---|---|---|---|---|
| Birds have wings | Declarative | I | I | Bayesian |
| Learn to ride a bike | Procedural | I, II | II | Bayesian variational |
| Solve a complex mathematical problem | Conditional | I, II | I, II | Bayesian variational |

Table 2: The process of knowledge generation according to the free energy framework of the brain for three different examples.

| | |
|---|---|
| Similarity matrix (mapping concepts to stimuli) | **A** |
| Transfer matrix (mapping previous concepts to new) | **B** |
| Prior distribution of stimuli | $C_\tau$ |
| Prior distribution of concepts (previous beliefs) | **D** |
| Accuracy parameter | $\gamma$ |
| Dirichlet parameters | $a, b, d$ |

$p(\mathbf{A}) = Dir(a), p(\mathbf{B}) = Dir(b), p(\mathbf{D}) = Dir(d)$

$p(\varphi_\tau | \theta_\tau) = Cat(\mathbf{A}), p(\theta_{\tau+1} | \theta_\tau, \pi) = Cat(\mathbf{B}_{\pi\tau})$

$p(\varphi_\tau) = Cat(\mathbf{C}), p(\theta_1) = Cat(\mathbf{D}), p(\pi) = \sigma(-\gamma \mathbf{G}(\pi))$

$\mathbf{G}(\pi) = \sum_\tau^T \mathbf{G}(\pi, \tau)$

Table 3: Variables and distribution functions of the generation model.

| | | |
|---|---|---|
| Perception | $q(\theta_\tau\|\pi) = arg_q min\,\mathbf{F}(\pi)$ <br> $\mathbf{F}(\pi) = E_q\left[-lnq(\tilde{\theta}\|\pi) - lnp(\tilde{\varphi},\tilde{\theta}\|\pi)\right]$ | From sensory stimuli to the generation of concepts |
| Planning | $q(\pi) = arg_q min\,\mathbf{F}$ <br> $\mathbf{F} = E_{q(\pi)}\left[\mathbf{F}(\pi) + lnq(\pi) + \mathbf{G}(\pi)\right]$ | From concepts to the selection of actions according to policies |
| Action | $a = arg_a max\left(q(\pi(\tau)=a)\right)$ | From actions to changing sensory stimuli |

**Table** 4: Perception, planning, and action in a process of generating concepts by minimizing free energy in the brain [37]

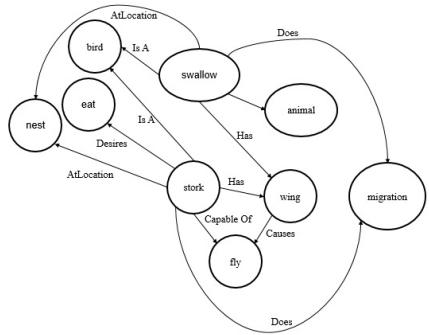

**Figure** 1: An example of a semantic network with 9 concepts (in the circles or ovals) and 12 relation links (arrows)

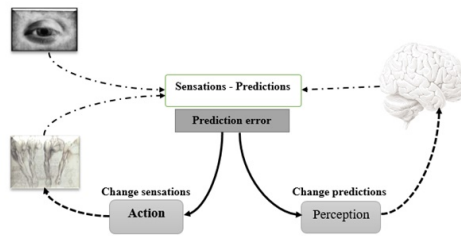

**Figure** 2: The process of learning concepts and updating the brain based on minimizing free energy and action-perception process. Prediction error can be reduced by changing predictions (perception) and sensations (action). Photo is recreated from [20] with permission.

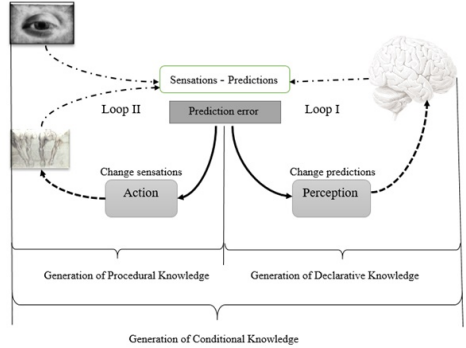

**Figure** 3: FEP brain model and knowledge generation
Loop I: generation of declarative knowledge (without active inference and independent of Loop II)
Loop II: generation of procedural knowledge (with active inference and connect to Loop I)
Combining two loops: generation of conditional knowledge

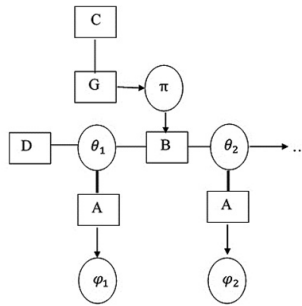

**Figure** 4: Generation of concepts in a Bayesian network according to the Markov decision-making process based on the principle of free energy. In this process, hidden concepts cannot be directly observed. The circles represent random variables, the comprise stimuli concepts and policies . Arrows indicate conditional probabilities. G, Expected Free Energy; D, initial concept prior; B, transition probabilities between hidden concepts; A, likelihood mapping from hidden concepts to stimuli.